\documentclass[11pt]{article}
\usepackage{fullpage,graphicx,psfrag,amsmath,amsfonts,verbatim}
\usepackage{xcolor}
\usepackage{amsthm}
\usepackage[small,bf]{caption}
\usepackage{authblk}

\usepackage{amsmath}
\usepackage{amssymb}
\usepackage{mathtools}
\usepackage{amsthm}
\usepackage{amsfonts}
\usepackage{algorithm}
\usepackage{algorithmic}

\usepackage{microtype}
\usepackage{graphicx}
\usepackage{booktabs} 
\usepackage{hyperref}
\usepackage{subcaption}

\allowdisplaybreaks

\title{\texttt{SkyRover}: A Modular Simulator for Cross-Domain Pathfinding}
\author[1]{Wenhui Ma\thanks{\texttt{wenhuima@stu.ecnu.edu.cn}}}
\author[2]{Wenhao Li\thanks{\texttt{whli@tongji.edu.cn}}}
\author[2]{Bo Jin\thanks{\texttt{bjin@tongji.edu.cn}}}
\author[3,4]{Changhong Lu\thanks{\texttt{chlu@math.ecnu.edu.cn}}}
\author[1,4]{Xiangfeng Wang\thanks{\texttt{xfwang@cs.ecnu.edu.cn}}}

\affil[1]{School of Computer Science and Technology, East China Normal University}
\affil[2]{School of Computer Science and Technology, Tongji University}
\affil[3]{School of Mathematical Sciences, East China Normal University}
\affil[4]{Key Laboratory of Mathematics and Engineering Applications, MoE}

\date{}
\begin{document}
\maketitle

\begin{abstract}

Unmanned Aerial Vehicles (UAVs) and Automated Guided Vehicles (AGVs) increasingly collaborate in logistics, surveillance, inspection tasks and etc. 
However, existing simulators often focus on a single domain, limiting cross-domain study. 
This paper presents the \texttt{SkyRover}, a modular simulator for UAV-AGV multi-agent pathfinding (MAPF). 
\texttt{SkyRover} supports realistic agent dynamics, configurable 3D environments, and convenient APIs for external solvers and learning methods. 
By unifying ground and aerial operations, it facilitates cross-domain algorithm design, testing, and benchmarking. 
Experiments highlight \texttt{SkyRover}’s capacity for efficient pathfinding and high-fidelity simulations in UAV-AGV coordination.
Project is available at \url{https://sites.google.com/view/mapf3d/home}.

\end{abstract}

\section{Introduction}\label{sec:intro}

The rapid growth of automation and artificial intelligence has significantly broadened unmanned systems’ application domains, especially in logistics and transportation. 
Companies like Amazon, JD, and Cainiao routinely deploy automated guided vehicles (AGVs) in large-scale warehouses\cite{qin2022jd,wurman2008coordinating,zhang2020learning}, while autonomous taxi services from Waymo and Baidu Apollo Go have begun trial operations globally\cite{sun2020scalability,wang2024recent}. 
Meanwhile, the low-altitude economy has gained momentum with drone-based package delivery, low-altitude tourism, and urban air mobility. 
Commercial drone delivery services by Amazon Prime Air and Meituan already enhance last-mile logistics\cite{dorling2016vehicle,engesser2023autonomous,sun2024uav}.


Alongside these developments, collaborative UAV–AGV systems are increasingly vital in freight transportation\cite{gao2020commanding}, search and rescue\cite{zhang2024air}, precision agriculture\cite{tokekar2016sensor}, and infrastructure inspection\cite{wu2020cooperative}, among other domains\cite{munasinghe2024comprehensive}. 
AGVs follow terrestrial routes, while UAVs operate in three-dimensional or restricted airspaces. 
Their synergy can boost operational capacity: UAVs excel in rapid transport of lightweight cargo, whereas AGVs handle heavier loads or site-specific tasks. 
However, integrating these systems demands careful scheduling and planning.


Multi-Agent Path Finding (MAPF)\cite{stern2019multi} is essential for collision-free, resource-efficient routing of large autonomous fleets. 
Although significant progress has been made for purely ground- or aerial-based MAPF\cite{alkazzi2024comprehensive,ma2022graph,salzman2020research,surynek2022problem}, new complexities arise when UAVs and AGVs share or intersect operational zones. 
Their heterogeneity (in motion patterns, energy consumption, and environmental constraints) further complicates joint planning; 
for instance, UAVs are more weather-sensitive, whereas AGVs often contend with dynamically changing indoor settings.


Robust simulation tools are key to designing, training, and testing MAPF algorithms for UAV–AGV scenarios, especially for data-intensive learning-based approaches.
Existing simulators handle single-domain movements effectively\cite{mavswarm2022,nguyen2019generalized,okumura2021iterative,skrynnik2024pogema,WangICAPS24mapf3d}, but few offer native support for UAV–AGV collaboration.


To fill this gap, we introduce \texttt{SkyRover}, a modular simulator for cross-domain pathfinding with UAV and AGV coordination. 
To our knowledge, it is the first environment providing a unified toolkit for UAV–AGV MAPF research. 
\texttt{SkyRover} allows users to build complex scenarios (e.g., warehouses and park environments) with adjustable fidelity and integrates high-fidelity physics models alongside user-friendly APIs for algorithmic testing and rapid prototyping. 
Through customizable and extensible interfaces, \texttt{SkyRover} supports a wide spectrum of use cases—from basic scheduling experiments to sophisticated learning-based methods.


\section{Related Work}

Extensive progress has been made in solving MAPF for single-domain vehicles\cite{ma2022graph,salzman2020research,surynek2022problem}. 
Previous simulators such as \cite{mavswarm2022,nguyen2019generalized,okumura2021iterative,skrynnik2024pogema,WangICAPS24mapf3d} have tackled various forms of agent-based simulation, yet few focus on integrated UAV-AGV scenarios. 
Moreover, many existing platforms concentrate on either 2D grids or simplified 3D representations, limiting the study of aerial and ground interactions. 
By contrast, \texttt{SkyRover} explicitly targets these cross-domain concerns, offering realistic physics, 3D occupancy grids, and unified APIs. 
To our knowledge, it is the first environment to natively support collaborative UAV-AGV MAPF under a single, modular framework.

\section{\texttt{SkyRover} Simulator}\label{sec:simulator}

\begin{figure}[htb!]
    \centering
    \includegraphics[width=\linewidth]{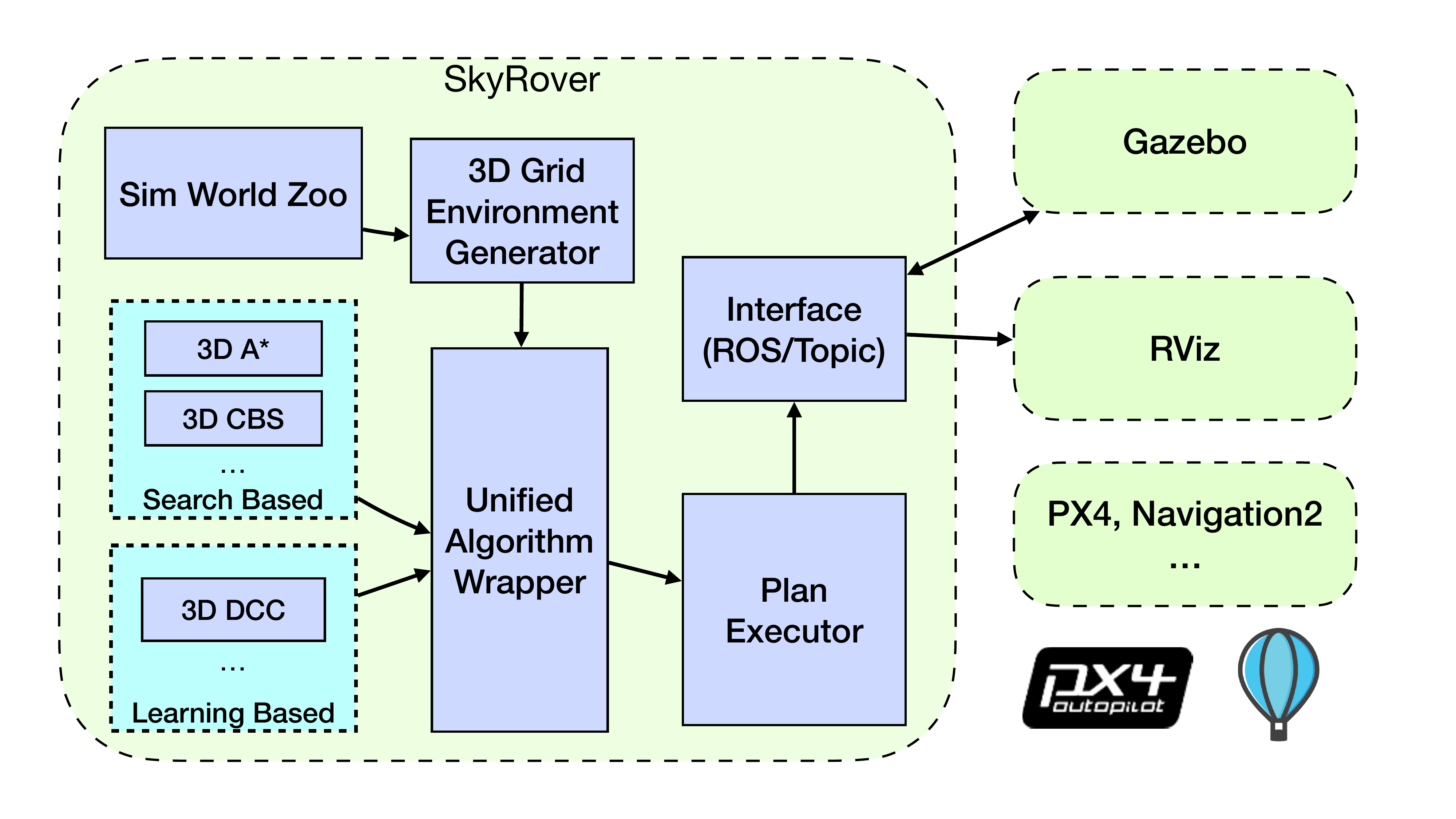}
    \caption{Main Architecture. \texttt{SkyRover} comprises multiple modules to support cross-domain MAPF.}
    \label{fig:skyrover}
\end{figure}

Following the motivations outlined in Section~\ref{sec:intro}, we present \texttt{SkyRover}, a modular simulator designed to address cross-domain pathfinding for UAV-AGV coordination. 
It aims to unify the modeling of ground and aerial vehicles in 3D grids while maintaining ease of use, modularity, and compatibility with third-party robotic frameworks. 
Figure~\ref{fig:skyrover} depicts its overall architecture, which comprises five main modules: 
1) the \textit{Sim World Zoo};
2) the \textit{3D Grid Generator};
3) the \textit{Unified Algorithm Wrapper};
4) the \textit{Plan Executor};
5) the \textit{System Interface}. 
This section details each module and explain how they collectively enable rapid experimentation and deployment of cross-domain MAPF solutions.

\subsection{\bf{Sim World Zoo}}

The \textit{Sim World Zoo} houses multiple Gazebo simulation worlds\cite{koenig2004design}, along with 3D models for UAVs and AGVs. 
Unlike simulators that rely solely on matrix-based grids, \texttt{SkyRover} offers more realistic environments by integrating detailed 3D worlds. 
Currently, \texttt{SkyRover} includes a warehouse and a park scenario, each featuring UAV and AGV models. 
These environments and virtual robots, sourced from \url{https://app.gazebosim.org/fuel/models}, can be further customized via the Gazebo UI. 
This design decision ensures that \texttt{SkyRover} users can scale from simple grid representations to complex settings that approximate real-world operational constraints.

\subsection{\bf{3D Grid Generator}}

Although \texttt{SkyRover} supports high-fidelity worlds, it also maintains a 3D grid representation for MAPF algorithms. 
The \textit{3D Grid Generator} automatically derives discrete map information from Gazebo environments. 
To achieve this, it employs a Gazebo plugin that creates a ROS2 service\cite{macenski2022robot} for capturing a 2D \texttt{.pgm} map and a 3D \texttt{.pcd} (point cloud) file. 
These files reflect the layout of objects and obstacles in the environment. 
Subsequently, the point cloud data is parsed to mark grid cells as free or occupied, generating a 3D occupancy grid suitable for both search- and learning-based algorithms.

\subsection{\bf{Unified Algorithm Wrapper}}

MAPF algorithms generally fall into either search-based or learning-based categories\cite{hart1968formal,sharon2015conflict,ma2021learning}. 
They also vary in their internal map structures, with some operating on matrices and others on more general graphs. 
To standardize these approaches, \texttt{SkyRover} provides a \textit{Unified Algorithm Wrapper} that abstracts the environment as a 3D grid with consistent interface calls, such as \textit{init}: initializes the environment, obstacle data, agent start states, and goal locations; \textit{step}: advances the simulation by one timestep, updating agent positions; and \textit{reset}: resets the simulation for new tasks or training episodes.
We provide 3D versions of popular MAPF algorithms, including 3D A*\cite{hart1968formal}, 3D CBS\cite{sharon2015conflict}, and a 3D extension of DCC\cite{ma2021learning}, ensuring that researchers and practitioners can rapidly evaluate both classical search and modern learning paradigms in a unified simulator.

\subsection{\bf{Plan Executor}}

High-level path planners often produce routes in discrete grids or graphs, but real robots need continuous control inputs. 
The \textit{Plan Executor} bridges this gap by translating each agent’s planned trajectory into commands interpretable by low-level controllers. 
The executor tracks each agent’s path, communicates with external control frameworks (e.g., PX4\cite{meier2015px4} or Navigation2\cite{nav2docs}), and updates agent positions in the simulator. 
This arrangement allows \texttt{SkyRover} to support both abstract, high-level pathfinding and more realistic, hardware-oriented simulations.


\subsection{\bf{System Interface}}

The \texttt{SkyRover} also offers multiple external interfaces to integrate with established robotics tools. 
Users can manipulate Gazebo models via Gazebo topics, visualize 3D occupancies with RViz\cite{kam2015rviz}, and command PX4-based drones through ROS2 topics and MicroXRCE\cite{microxrcedds}. 
This flexibility supports varied applications, ranging from hardware-in-the-loop testing to large-scale simulation. 
Ultimately, \texttt{SkyRover} allows researchers and practitioners to incorporate realistic dynamics, robust visualization, and real-time interactivity into customized UAV-AGV cooperative scenarios.

\section{Experiment}\label{sec:exp}

In this section, we demonstrate how \texttt{SkyRover} supports diverse experimental setups. 
We showcase the simulator using two distinct Gazebo worlds: a warehouse (Figure~\ref{fig:detail_warehouse}) and a park (Figure~\ref{fig:gazebo_building}). 
These worlds serve as representative testbeds, featuring varied layouts that allow us to highlight the simulator’s 3D mapping, pathfinding, and visualization capabilities. We define two typical AGV-UAV interaction tasks: 
\begin{itemize}
    \item \textbf{Inventory Scanning Task}: The AGV transports cargo to point A, where a UAV hovers above the AGV to scan and inventory the cargo. After the scanning process is completed, the AGV continues transporting the cargo to point B;
    \item \textbf{Aerial Cargo Transfer Task}: The AGV transports cargo to point A, where a UAV arrives and hovers above the AGV to pick up the cargo. The UAV then lifts the cargo and transports it to an elevated point B.
\end{itemize}

These tasks demonstrate the seamless coordination between AGVs and UAVs in different scenarios, showcasing the effectiveness of \texttt{SkyRover} in simulating complex robotic interactions.

\subsection{\bf{Environment Setup}}

\begin{figure}[htb!]
\centering
    \begin{subfigure}[t]{0.48\linewidth}
        \centering
        \includegraphics[width=\linewidth]{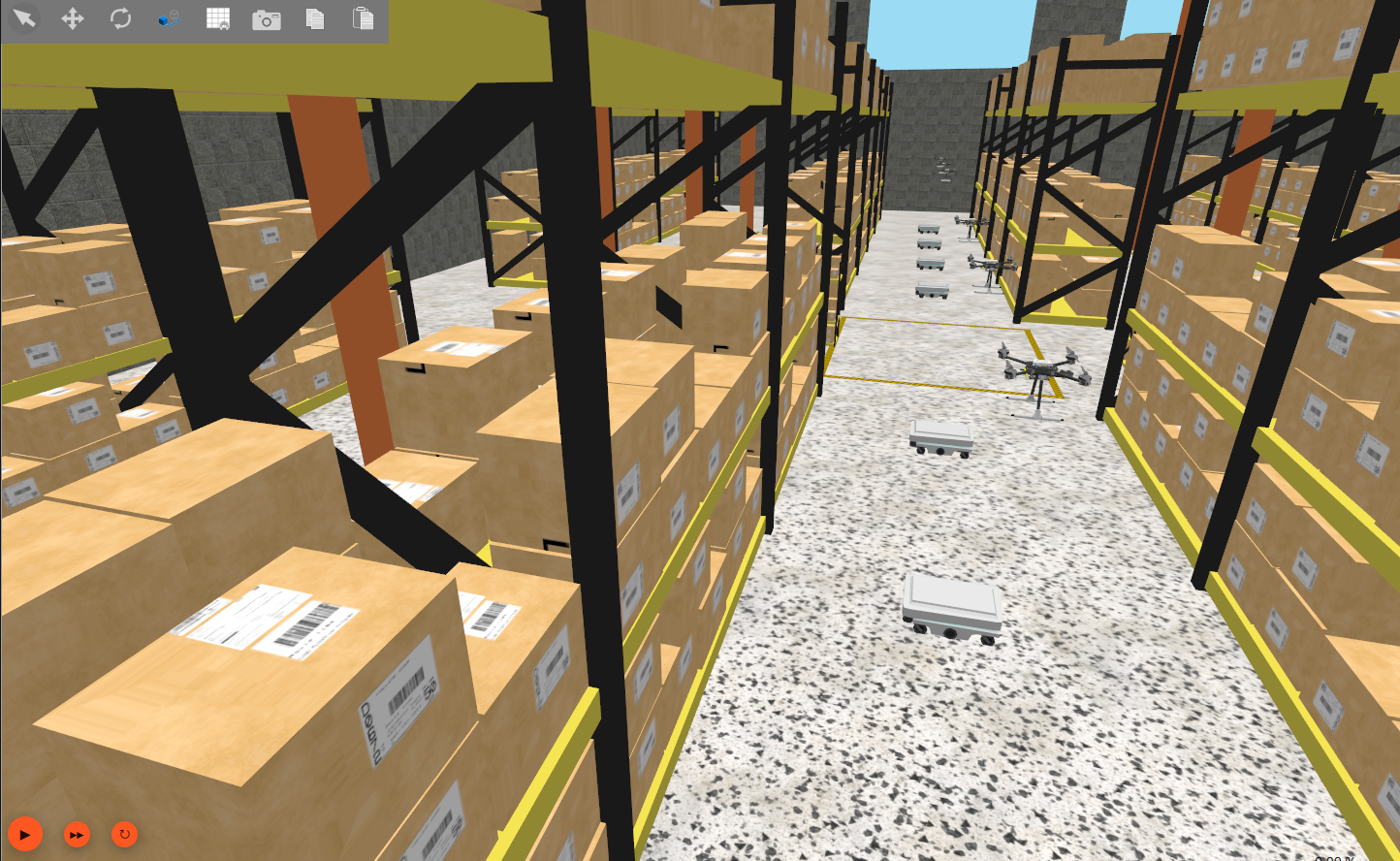}
        \caption{}
        \label{fig:detail_warehouse}
    \end{subfigure}
    \begin{subfigure}[t]{0.48\linewidth}
        \centering
        \includegraphics[width=\linewidth]{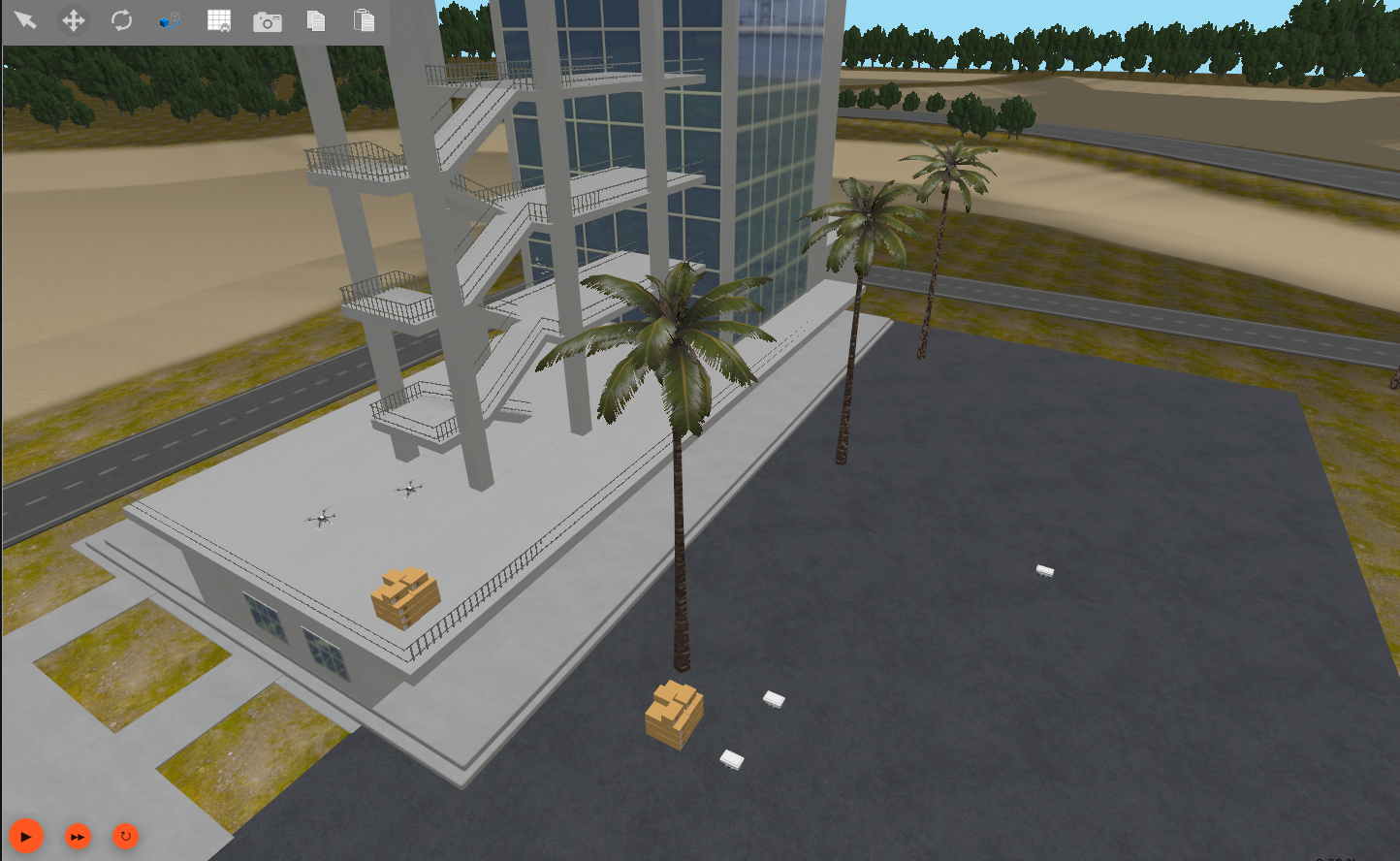}
        \caption{}
        \label{fig:gazebo_building}
    \end{subfigure}
    \begin{subfigure}[t]{0.48\linewidth}
        \centering  
        \includegraphics[width=\linewidth]{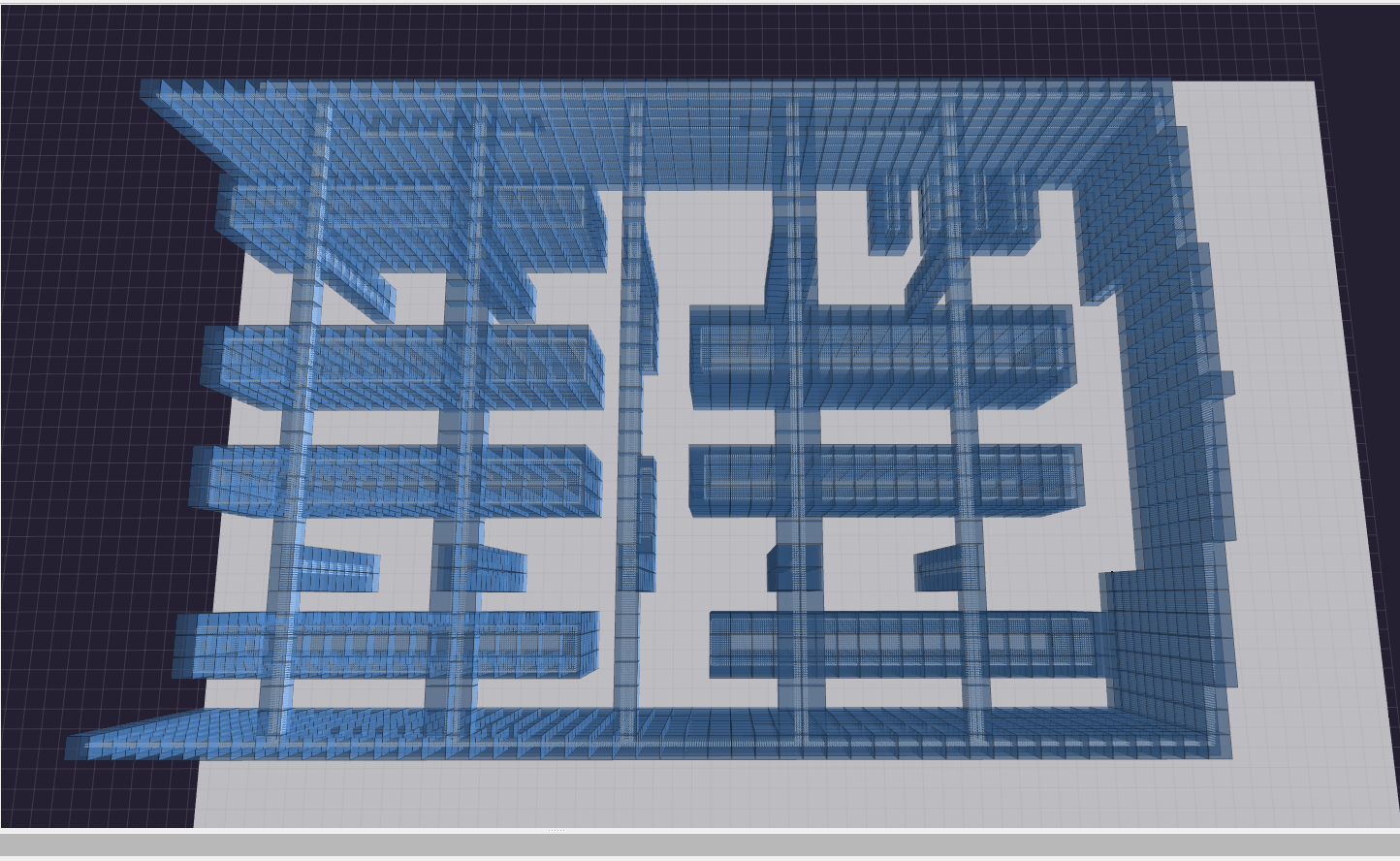}
        \caption{}
        \label{fig:rviz_warehouse}
    \end{subfigure}
    \begin{subfigure}[t]{0.48\linewidth}
        \centering
        \includegraphics[width=\linewidth]{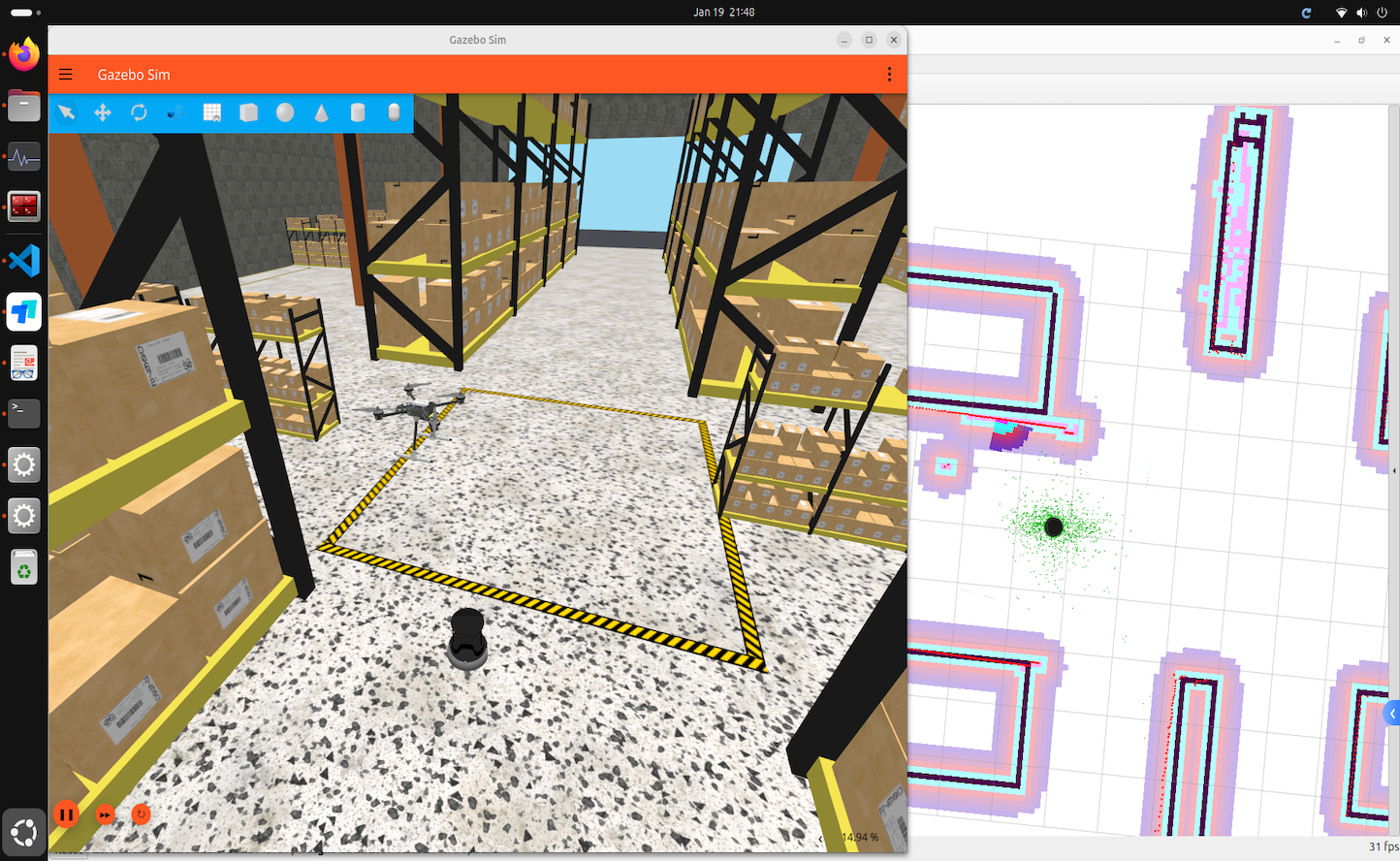}
        \caption{}
        \label{fig:low-level}
    \end{subfigure}
    \caption{(a) The warehouse Gazebo world, featuring multiple Holybro X500 drones and delivery AGVs; (b) The park scenario, offering more open space for UAV operations; (c) The 3D occupancy grid in RViz. Each dark cell has point cloud data and is thus considered an obstacle; (d) Example of integrating \texttt{SkyRover} with hardware-oriented controllers. PX4 executes drone flight commands and Navigation2 governs the TurtleBot.}
    \label{fig:scenarios}
\end{figure}

We begin by loading the chosen Gazebo world and generating a 3D occupancy grid using the gazebo-map-creator plugin\cite{arshad2022}. 
This plugin extracts a point cloud of the environment, which is converted into a $0$--$1$ grid where each cell spans \(1\)\,meter. 
Cells containing point cloud particles are treated as obstacles and labeled ``1,'' while free cells remain ``0.'' 
The resulting 3D grid is published to RViz for real-time visualization (Figure~\ref{fig:rviz_warehouse}). 





\subsection{\bf{3D MAPF Examples}}

After generating the occupancy grid, we deploy $6$ Holybro X500 drones and $16$ AGVs. 
We assign each agent a unique start and goal location in the warehouse world. 
Via the \textit{init} interface of algorithm wrapper, we load the 3D-A* and 3D-CBS solvers. 
During initialization, these solvers compute complete, conflict-free paths for all agents. 
At each timestep, the \textit{step} function moves every agent to its next waypoint, ensuring collision-free navigation. 

We also implement a 3D version of the learning-based method, DCC\cite{ma2021learning}. 
Here, we adapt the original 2D convolution layers to 3D, and employ curriculum learning to train for \(70{,}000\) episodes in a random \(40\times40\times40\) grid. 
This trained model reaches $100\%$ success under test conditions with sixteen agents. 
Unlike search-based algorithms, which plan entire routes beforehand, the learning-based approach invokes the \textit{step} function to infer actions online after loading the model file. 
This approach suits dynamic environments where complete pre-planning is less feasible.

\subsection{\bf{Motion Control Integration}}

For many studies, abstract position updates are sufficient for benchmarking algorithmic performance. 
In these cases, we simply invoke Gazebo’s model-position service to teleport each agent, bypassing detailed dynamics. 
However, \texttt{SkyRover} also supports low-level motion control through PX4\cite{meier2015px4} and Navigation2\cite{nav2docs}. 
Figure~\ref{fig:low-level} shows an example involving a drone controlled by PX4 and a TurtleBot commanded by Navigation2. 
These finer-grained simulations accurately capture kinematic and dynamic constraints, which are essential for hardware-in-the-loop testing. 
Although more computationally intensive, these setups are valuable for research on real-time control, multi-robot coordination, and safety validation. 
Users can choose the approach—abstract or low-level—based on the complexity of their experiments and available computing resources.

\subsection{\bf{Preliminary Results and Comparison}}

We conduct all experiments on Ubuntu 24.04 with ROS 2 Jazzy, Gazebo Harmonic, and the main branch of PX4, using a PC with an Intel i7 CPU and 32\,GB RAM. 
Table~\ref{tab:sim_results} presents preliminary performance metrics comparing 3D variants of A*, CBS and DCC on the warehouse world when $22$ total agents ($6$ drones, $16$ AGVs) must reach randomly assigned goals. 
\textbf{Computation time} measures the total time to compute or infer a path before the first move, while \textbf{Success rate} indicates the percentage of agents reaching their target without collision in a single run. 

\begin{table}[ht]
\centering
\caption{Preliminary Results in the Warehouse (22 Agents).}
\label{tab:sim_results}
\resizebox{.7\linewidth}{!}{%
\begin{tabular}{lcc}
\hline
\textbf{Algorithm} & \textbf{Comp. Time (s)} & \textbf{Success Rate ($\%$)}\\
\hline
3D-A*   & 54.7 & 100 \\
3D-CBS & 92.4 & 100 \\
3D-DCC (Well-trained) & 0.6 & 100 \\
\hline
\end{tabular}%
}
\end{table}

These initial comparisons highlight \texttt{SkyRover}’s ability to benchmark MAPF algorithms in a unified and consistent framework. 
Future work could involve more in-depth evaluations, including scaling to larger agent teams and investigating runtime on different hardware setups.

\section{Conclusion and Future Directions}\label{sec:conclusion}

In this paper, we introduce \texttt{SkyRover}, a modular simulator paving the way for integrated UAV-AGV MAPF in 3D environments. 
It combines realistic Gazebo worlds, a robust 3D grid generator, unified wrappers for classical and learning-based algorithms, and seamless integration with external robotics software. 
Experiments in the warehouse and park worlds confirmed its flexibility for discrete pathfinding and low-level control simulations.

\paragraph{Limitations}
Current limitations include partial modeling of real-world effects (e.g., weather, sensor noise) and the computational load of large-scale simulations. 
Learning-based methods also require extensive training data, and the hyperparameter optimization might be time-intensive.

\paragraph{Future Directions}
Going forward, we plan to integrate more realistic physics (wind perturbations, complex friction models), advanced sensor types (LiDAR, radar), and dynamic obstacle handling. 
Collaborations with urban air traffic simulators would further expand scenario possibilities, such as multi-lane sky corridors for UAV delivery. 
Additionally, a systematic approach to large-scale distributed training could support RL methods that tackle hundreds of agents in real time.

\newcommand{\etalchar}[1]{$^{#1}$}

\end{document}